\documentclass[letterpaper, 10 pt, conference]{ieeeconf}  

\IEEEoverridecommandlockouts                              

\usepackage{graphicx} 
\usepackage{amsmath} 
\usepackage{amssymb}  
\usepackage{booktabs}
\usepackage{tabularx}
\usepackage{pifont}
\usepackage[table]{xcolor}
\definecolor{projectblue}{HTML}{246BA0}
\usepackage{xspace}
\usepackage{flushend}
\newcommand{\methodname}{Rolling-WAM\xspace}

\newcommand{\gOnePiDog}{55.0}
\newcommand{\gOnePiPlate}{70.0}
\newcommand{\gOnePiPour}{60.0}
\newcommand{\gOnePiAverage}{61.7}
\newcommand{\gOneGrootDog}{75.0}
\newcommand{\gOneGrootPlate}{65.0}
\newcommand{\gOneGrootPour}{60.0}
\newcommand{\gOneGrootAverage}{66.7}
\newcommand{\gOneJointDog}{70.0}
\newcommand{\gOneJointPlate}{100.0}
\newcommand{\gOneJointPour}{65.0}
\newcommand{\gOneJointAverage}{78.3}
\newcommand{\gOneFastDog}{85.0}
\newcommand{\gOneFastPlate}{80.0}
\newcommand{\gOneFastPour}{60.0}
\newcommand{\gOneFastAverage}{75.0}
\newcommand{\gOneRollingDog}{85.0}
\newcommand{\gOneRollingPlate}{100.0}
\newcommand{\gOneRollingPour}{70.0}
\newcommand{\gOneRollingAverage}{85.0}

\newcommand{\cmark}{\ding{51}}
\newcommand{\xmark}{\ding{55}}

\usepackage[hidelinks]{hyperref}

\makeatletter
\let\ICRA@makecaption\@makecaption
\long\def\@makecaption#1#2{%
  \begingroup
    \let\footnotesize\normalsize
    \ifx\@captype\@IEEEtablestring%
      \setbox\@tempboxa\hbox{\footnotesize #1: #2}%
      \ifdim\wd\@tempboxa>\hsize
        \parbox[t]{\hsize}{\footnotesize\noindent #1: #2}%
      \else
        \hbox to\hsize{\footnotesize\hfil\box\@tempboxa\hfil}%
      \fi
      \@IEEEtablecaptionsepspace
    \else
      \ICRA@makecaption{#1}{#2}%
    \fi
  \endgroup
}
\makeatother

\title{\LARGE \bf
\methodname: World Action Models with Rolling Imagination
}

\makeatletter
\renewcommand{\@IEEEauthorblockAtopspace}{8pt}
\makeatother

\author{%
\authorblockN{%
Yinghua Zhou\textsuperscript{1,2,*}\quad
Junjie Ye\textsuperscript{1,*}\quad
Yiqi Zhao\textsuperscript{1}\quad
Hao Dong\textsuperscript{1}\quad
Celina Shiyu Wang\textsuperscript{1}\quad
Ruohai Ge\textsuperscript{1}\\
Tingyi Yang\textsuperscript{1,3}\quad
Basile Van Hoorick\textsuperscript{4}\quad
Gaurav Sukhatme\textsuperscript{1}\quad
Vitor Guizilini\textsuperscript{4,\textdagger}\quad
Yue Wang\textsuperscript{1,\textdagger}}
\authorblockA{%
\textsuperscript{1}University of Southern California\quad
\textsuperscript{2}Brown University\quad
\textsuperscript{3}Fudan University\quad
\textsuperscript{4}Toyota Research Institute\\
\textsuperscript{*}Equal contribution\quad
\textsuperscript{\textdagger}Equal advising\\[4pt]
Project page: \textcolor{projectblue}{\url{https://rolling-wam.github.io/}}}
}
\hypersetup{pdfauthor={Yinghua Zhou; Junjie Ye; Yiqi Zhao; Hao Dong; Celina Shiyu Wang; Ruohai Ge; Tingyi Yang; Basile Van Hoorick; Gaurav Sukhatme; Vitor Guizilini; Yue Wang}}

\makeatletter
\IEEEaftertitletext{%
  \begin{minipage}{\textwidth}
    \centering
    \includegraphics[width=\linewidth]{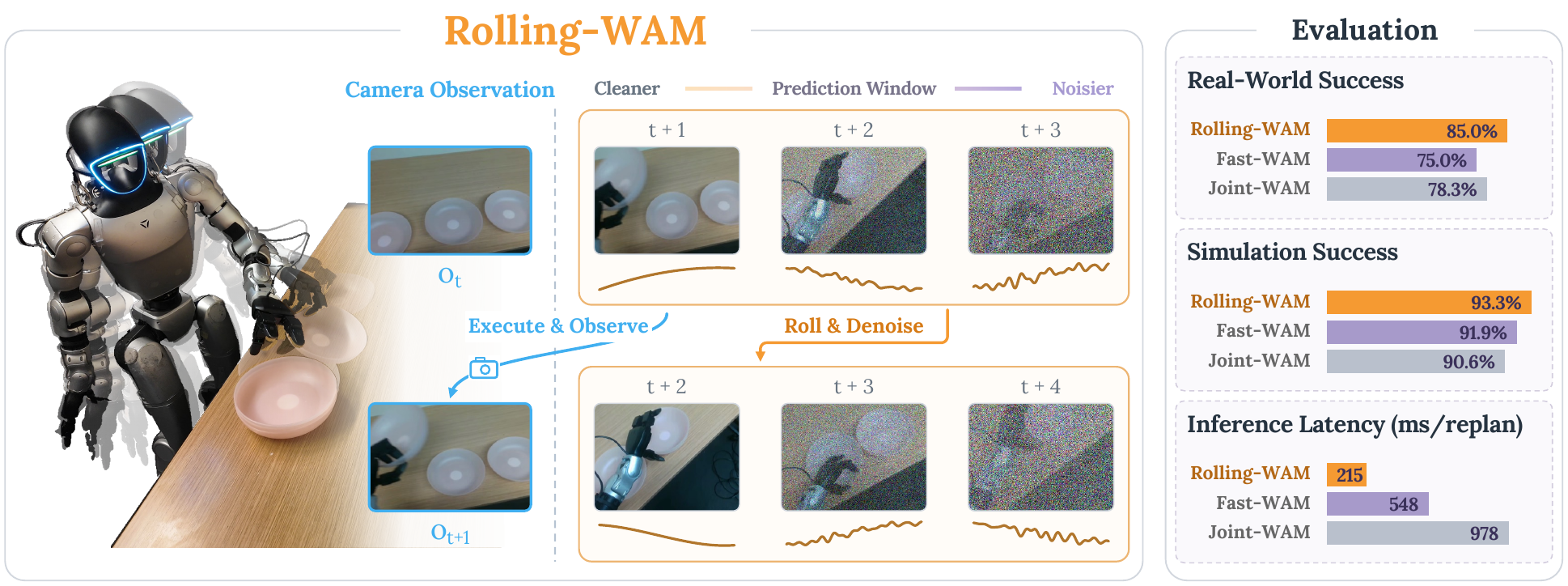}
    \def\@captype{figure}
    \vspace{-15pt}
    \caption{\textbf{\methodname maintains a rolling window of video-action predictions at staggered noise levels.}
    For simplicity, each frame represents one chunk, and the traces denote actions.
    After executing the first action chunk, the robot acquires a new camera observation,
    shifts the window, and denoises the retained future chunks together with a newly appended chunk initialized with Gaussian noise.
    This approach distributes the denoising cost across control cycles, achieving competitive task performance with substantially faster inference.}
    \label{fig:teaser}
  \end{minipage}%
  \vspace{\baselineskip}%
}
\makeatother

\begin{document}

\maketitle
\thispagestyle{empty}
\pagestyle{empty}

\begin{abstract}
World Action Models (WAMs) couple action generation with future visual prediction for robotic manipulation.
However, completing the joint video-action denoising process at each replanning cycle incurs substantial latency, delaying action updates and limiting closed-loop responsiveness.
We present \methodname, a formulation that distributes joint denoising across successive replanning cycles. 
Our method maintains a sliding window of video-action chunks at staggered noise levels. 
At each step, a rolling noise schedule fully denoises the imminent action chunk for execution, 
while partially refining farther-future chunks. 
As the window advances with new camera observations, the retained future chunks continue their denoising process. 
This distributes the computational cost over time while carrying an evolving visual-action context across chunk boundaries. 
Evaluations on LIBERO, RoboTwin, and a real-world Unitree G1 humanoid show that \methodname achieves competitive manipulation performance. 
By removing the need to denoise the entire prediction horizon from scratch, it delivers a $4.5\times$ steady-state replanning speedup over standard joint WAMs.
\end{abstract}

\section{Introduction}


World Action Models (WAMs) have recently advanced robotic
manipulation by jointly predicting robot actions and future
observations~\cite{lingbotva,dreamzero,cosmos3}.
This joint modeling provides visual context for action
generation, helping the policy anticipate how the scene
will evolve.
However, deployment in dynamic environments requires
closed-loop control: the robot must repeatedly replan from
the latest observations to correct execution errors and
respond to environmental changes.

Frequent replanning exposes a computational bottleneck in
standard WAMs. Joint denoising samplers process the entire
prediction horizon from pure noise to clean data at each
replanning cycle.
Each denoising step processes the entire video-action sequence, making repeated model evaluations computationally
expensive.
The resulting inference latency delays action updates, limiting closed-loop responsiveness. Some methods achieve faster inference by omitting future-video
generation at test time~\cite{fastwam}, at the cost of losing
explicit visual predictions that could inform action generation.


We argue that denoising can be organized more efficiently
across replanning cycles.
Conventional chunk-based samplers~\cite{dreamzero,fastwam}
fully denoise a new prediction horizon at each replan,
even when receding-horizon control executes only an
initial portion.
The unexecuted tail can inform current actions but is
discarded rather than reused across replans.
Retaining and progressively refining future video--action
chunks across cycles can preserve predictive context and
reuse denoising computation, reducing the sequential steps
needed to produce the next executable chunk.


We therefore introduce \methodname, a formulation that
inherits this intuition by distributing denoising across
successive replanning cycles.
Building on the concept of rolling diffusion for sequence
generation~\cite{rollingdiffusion,rollingforcing},
\methodname maintains a sliding window of aligned
video-action chunks with progressively higher noise
levels toward the future (Fig.~\ref{fig:teaser}).
At each replanning cycle, only the imminent action chunk is
fully denoised for execution, while farther-future chunks
remain partially denoised.
This resembles the intuition behind human planning:
near-term actions are concrete, while more distant plans
remain provisional and are refined as new observations arrive.


After the first action chunk is executed, the window advances,
retaining the remaining predictions and appending a new chunk
initialized with Gaussian noise.
The retained predictions are then jointly refined with the
new chunk, conditioned on the newly acquired camera observation.
As chunks move through the window, their denoising steps
are distributed across replanning cycles.
Each steady-state cycle therefore requires only a fraction
of the total steps to produce the next executable chunk.
We train the model on matching noise profiles, with action
tokens attending to partially denoised visual futures
across the window.


We evaluate \methodname on simulation benchmarks, including LIBERO~\cite{libero} and RoboTwin~\cite{robotwin}, as well as three real-world humanoid manipulation tasks on the Unitree G1. 
Our experiments demonstrate that \methodname retains the benefits of visual imagination while significantly reducing computational overhead. 
It achieves a 98.1\% average success rate on LIBERO and 93.3\% on RoboTwin, remaining competitive with or exceeding state-of-the-art WAMs. 

Our primary contribution is a rolling formulation for joint video-action models that distributes denoising across successive replanning cycles. 
We demonstrate that this formulation achieves a $4.5\times$ inference speedup over standard WAMs while maintaining competitive manipulation performance.

\begin{figure*}[t]
    \centering
    \includegraphics[width=\textwidth]{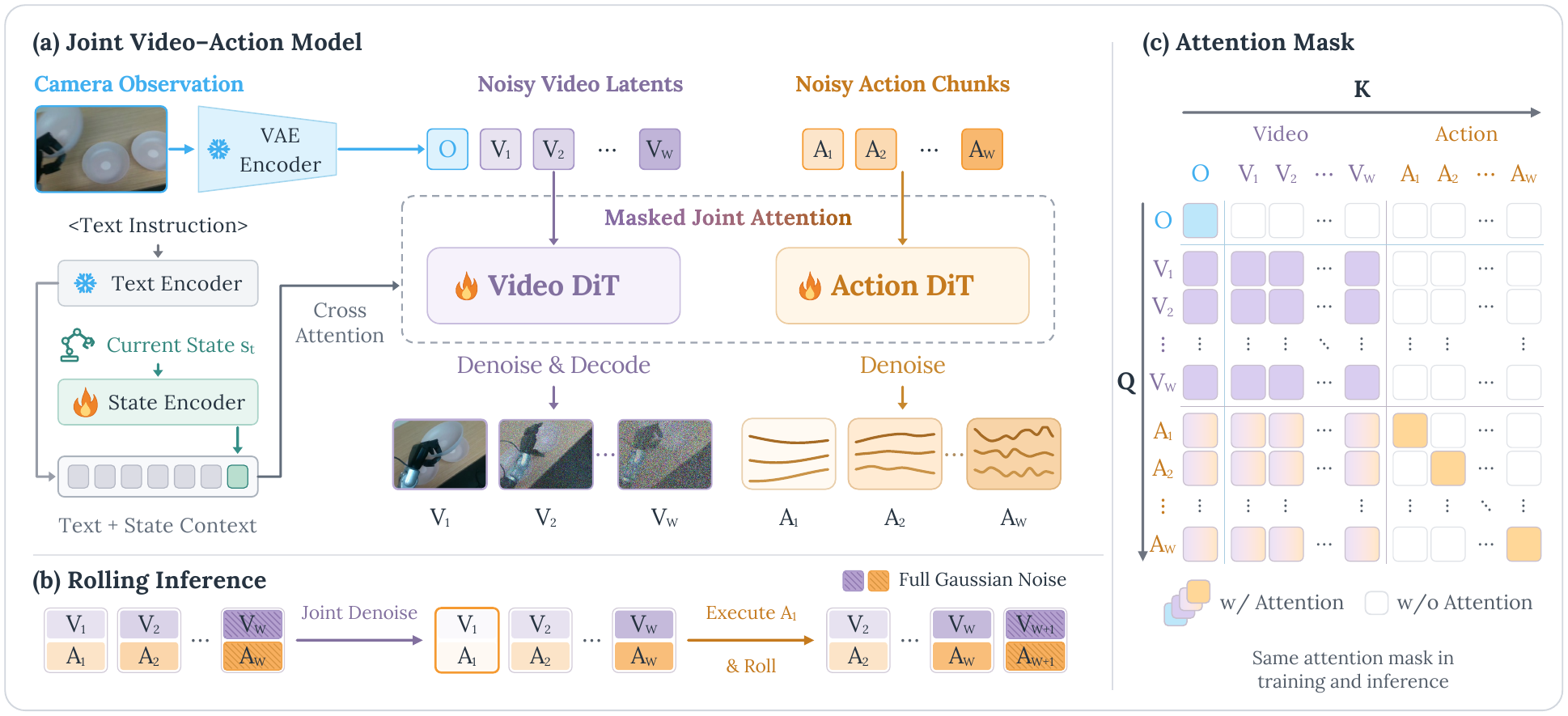}
    \vspace{-12pt}
    \caption{\textbf{\methodname framework.}
    (a) Video and action experts jointly predict flow velocities using masked joint attention and cross-attention to text and state context, conditioned on the current camera observation and per-chunk noise levels shared by video and actions.
    (b) Rolling inference denoises the window, executes the first action chunk, and shifts the retained future chunks while appending a new chunk initialized with Gaussian noise.
    (c) The same attention mask is used during training and inference.}
    \vspace{-12pt}
    \label{fig:framework}
\end{figure*}

\section{Related Work}

\subsection{World Action Models for Robotic Manipulation}
World Action Models (WAMs) augment action generation with future visual prediction. 
Early approaches infer actions from language-conditioned video plans~\cite{unipi}. 
Recent methods integrate visual prediction and action generation using pretrained video models~\cite{lingbotva,dreamzero,cosmos-policy}, often incorporating multimodal understanding~\cite{motus,cosmos3} or exploring native causal video-action pretraining~\cite{lingbotva2}. 
Alternative designs learn predictive latent representations for action generation to avoid iterative future-video denoising at deployment~\cite{internvlaa15,omega0}. 
For joint WAMs, however, iterative video-action denoising remains computationally costly, motivating more efficient model designs and sampling strategies.

\subsection{Efficient Inference for World Action Models}
Accelerating WAM inference typically involves reducing the computational burden of future prediction. 
Some methods remove future-video generation entirely at deployment~\cite{fastwam} or extract future context in a single video-expert pass to cache for action denoising~\cite{fasterwam}. 
Other architectural optimizations include reducing model size, visual tokens, and video denoising steps~\cite{efficientwam}, or accelerating sampling through modality-aware consistency distillation~\cite{flashwam}. 
Other strategies focus on context reuse, either by recycling visual features after partial joint denoising~\cite{motubrain} or by sharing asynchronously refreshed visual context across action updates~\cite{ahawam}. 
Complementarily, real-time chunking overlaps inference with execution and uses action inpainting to align successive chunks~\cite{rtc, psi0}. 
Instead of removing or caching future predictions, \methodname distributes the joint video-action denoising process across replanning cycles, progressively refining near-term and future predictions within a sliding window.

\subsection{Rolling Diffusion for Sequence Generation}
Rolling Diffusion~\cite{rollingdiffusion} introduces a sliding denoising window with progressively higher noise levels toward the future. 
This concept has been extended with independent token noise levels for flexible sampling schedules~\cite{diffusionforcing} and bidirectional denoising with rollout-based distillation for autoregressive long-video generation~\cite{rollingforcing}. 
These works motivate the rolling formulation of \methodname.

In robotic manipulation, Streaming Diffusion Policy~\cite{sdp} and RNR-DP~\cite{rnr-dp} maintain partially denoised action buffers to accelerate action generation. 
However, these methods focus exclusively on action-only denoising and evaluate on task-specific policies in relatively small-scale settings. 
We extend this rolling mechanism to couple world modeling with action generation, enabling action tokens to attend to partially denoised visual futures, and evaluate our approach across broader multitask settings.

\section{Method}
\label{sec:method}

This section presents \methodname, a joint video-action model for low-latency closed-loop robotic control. 
We first formulate the visual-action modeling problem and identify the computational bottleneck of standard joint denoising (Sec.~\ref{sec:formulation}). 
We then introduce a rolling denoising mechanism that maintains a sliding window of predictions at staggered noise levels (Sec.~\ref{sec:rolling}). 
Next, we describe the sampling and execution process that distributes computation across control cycles (Sec.~\ref{sec:sampling}). 
Finally, we detail the model architecture and the training objective (Sec.~\ref{sec:model_training}).

\subsection{Problem Formulation}
\label{sec:formulation}

We consider language-conditioned robotic manipulation from visual observations and proprioception. 
At time $t$, the policy receives an observation $o_t$, robot state $s_t$, and language instruction $\ell$. 
It predicts an action sequence $\mathbf{a}_{t:t+H-1}$ over a horizon of $H$ steps. 
World Action Models (WAMs) augment this action generation with future visual prediction~\cite{lingbotva,dreamzero}. 
Let $\mathbf{v}_{t+1:t+H}$ denote the future video latents covering the corresponding physical interval at the video sampling rate. 
The joint visual-action modeling problem is formulated as learning the distribution:
\begin{equation}
    p_\theta\!\left(
        \mathbf{a}_{t:t+H-1},\mathbf{v}_{t+1:t+H}\mid c_t
    \right),
    \qquad c_t=(o_t,s_t,\ell).
    \label{eq:wam_formulation}
\end{equation}
This distribution can be modeled jointly or factorized with action generation conditioned on future visual prediction. 

Deploying WAMs in a closed loop requires multiple denoising steps within each replanning cycle before an action chunk is executed. 
This iterative process incurs a high computational cost. 
It increases inference latency and limits the responsiveness of the controller to new observations. 

As shown in Fig.~\ref{fig:framework}, \methodname addresses this bottleneck by distributing the denoising computation across successive replanning cycles. 
It maintains a sliding window that jointly refines the next action chunk and its future continuation. 
This approach completes near-term predictions while leaving longer-term chunks partially denoised at staggered noise levels. 
As the window advances with execution, the retained predictions continue to evolve under new observations. 
This mechanism carries predictive context across chunk boundaries, allowing successive action chunks to be generated with a shared, evolving visual-action future. 

\subsection{Rolling Video-Action Denoising}
\label{sec:rolling}

We partition the prediction window into $W$ temporally aligned chunks $\mathbf{X}_{1:W}$, where $\mathbf{X}_j=(\mathbf{v}^{(j)},\mathbf{a}^{(j)})$ denotes the $j$-th video-action chunk. 
Each action chunk $\mathbf{a}^{(j)}=\mathbf{a}_{t+(j-1)K:t+jK-1}$ contains $K$ actions. 
Its corresponding video chunk $\mathbf{v}^{(j)}$ covers the same physical interval at the designed video sampling rate. 
The total prediction horizon is therefore $H=WK$. 
Execution advances by one chunk at a time. 
During each replanning cycle, the window is processed jointly, but only the nearest chunk reaches completion.

Let $\tau\in[0,1]$ parameterize the denoising phase of a single replanning cycle, decreasing from $1$ at the start to $0$ at completion. 
Let $\sigma(\cdot)$ denote a monotonically increasing base noise schedule, with $\sigma(1)=1$ corresponding to Gaussian noise and $\sigma(0)=0$ to clean data. 
We define two operational modes to establish and maintain the rolling window.

\paragraph{Rolling mode}
The first chunk must become clean at the end of a replanning cycle. 
Each remaining chunk must be ready to occupy the preceding position in the window. 
We assign chunk $j$ the noise level~\cite{rollingdiffusion}:
\begin{equation}
    \sigma_j^{\mathrm{rolling}}(\tau)
    =\sigma\!\left(\frac{j-1+\tau}{W}\right),
    \qquad j=1,\ldots,W.
    \label{eq:rolling_schedule}
\end{equation}
As $\tau$ decreases from $1$ to $0$, chunk $j$ moves from noise level $\sigma(j/W)$ to $\sigma((j-1)/W)$. 
The first chunk becomes clean, while later chunks remain at progressively higher noise levels. 
Crucially,
\begin{equation}
    \sigma_{j+1}^{\mathrm{rolling}}(0)
    =\sigma_j^{\mathrm{rolling}}(1),
    \qquad j=1,\ldots,W-1.
    \label{eq:rolling_alignment}
\end{equation}
After the first chunk is removed, every retained chunk is already at the starting noise level for its new position. 
We append a new chunk initialized with Gaussian noise at $\sigma=1$, restoring the configuration for the next replanning cycle.

\paragraph{Initialization mode}
At the beginning of an episode, partially denoised predictions are unavailable. 
We initialize the entire window from Gaussian noise and use the schedule:
\begin{equation}
    \sigma_j^{\mathrm{init}}(\tau)
    =\sigma\!\left(
        \min\left\{1,\;\tau+\frac{j-1}{W}\right\}
    \right).
    \label{eq:initialization_schedule}
\end{equation}
All chunks start at a noise level $\sigma=1$. 
Nearer chunks begin denoising earlier, while more distant chunks remain at pure noise until their refinement begins. 
At completion, $\sigma_j^{\mathrm{init}}(0)=\sigma_j^{\mathrm{rolling}}(0)$. 
The first chunk is executable, and the remaining predictions enter rolling mode after the window shifts. 

Together, these two modes allow each chunk to begin as a distant prediction and receive further refinement as it approaches execution. 
Video and actions within each chunk share the same denoising steps.

\subsection{Sampling and Execution}
\label{sec:sampling}

We allocate $N$ total denoising steps per chunk. 
We choose $N$ as a multiple of $W$ to ensure an even distribution across rolling cycles. 
Initialization uses $\Delta\tau=-1/N$ at each denoising step, requiring $N$ steps to reach rolling mode. 
Subsequent replanning cycles require only $N/W$ steps to produce the next executable action chunk, with $\Delta\tau=-W/N$ at each step.

At each denoising step, the model jointly updates the video and action predictions throughout the window. 
Let $\widetilde{\mathbf{X}}_j$ denote the current denoising state of chunk $j$. 
Given the predicted flow velocity $f_{\theta,j}$, the Euler update is:
\begin{equation}
    \widetilde{\mathbf{X}}_j
    \leftarrow
    \widetilde{\mathbf{X}}_j
    +
    \bigl[
        \boldsymbol{\sigma}_j(\tau+\Delta\tau)
        -\boldsymbol{\sigma}_j(\tau)
    \bigr]
    f_{\theta,j}\!\left(
        \widetilde{\mathbf{X}}_{1:W},
        \boldsymbol{\sigma};c_t
    \right).
    \label{eq:euler_update}
\end{equation}

Once the first chunk is clean, the robot executes its $K$ actions. 
The window then slides over. 
It removes the executed chunk, retains the remaining predictions, and appends a new chunk initialized with Gaussian noise (Fig.~\ref{fig:framework}b). 
The robot acquires the latest camera observation and state to condition the next replanning cycle.

Each chunk traverses all $N$ denoising steps before execution, but its computation is distributed across multiple replanning cycles. 
This mechanism reduces the number of denoising steps needed to produce the next executable chunk, while preserving an evolving visual-action prediction across chunk boundaries. 
Consequently, larger windows can lower steady-state inference latency, assuming fixed $N$ and sufficient GPU parallelism to process the extended prediction window. 
This approach is orthogonal to other efficiency techniques~\cite{fasterwam,efficientwam} for WAMs and can be combined to further reduce inference latency.

\subsection{Model Architecture and Training}
\label{sec:model_training}

As shown in Fig.~\ref{fig:framework}(a), \methodname pairs a pretrained video Diffusion Transformer~\cite{wan2025,dit}, whose visual priors transfer well to robot manipulation videos~\cite{robodream, anchordream}, with a lightweight action Transformer in a Mixture-of-Transformers (MoT) architecture~\cite{fastwam,mixtureoftransformers}. 
The video expert processes VAE latents of the current observation and noisy future video. 
The action expert processes projected noisy actions. 
Masked joint attention couples the experts at each layer, allowing action generation to draw on pretrained visual dynamics. 
Language and proprioceptive embeddings condition both experts through cross-attention.

To support rolling denoising, video and action tokens are modulated by their respective chunk-wise noise levels. 
This allows predictions at different stages of refinement to be processed jointly. 
The attention mask (Fig.~\ref{fig:framework}c) allows every action chunk to attend to visual features across the entire prediction window. 
Direct action-to-action attention is restricted to the same chunk, and video tokens do not attend to action tokens. 
Each action chunk therefore draws on a shared, evolving visual prediction that extends beyond its own execution interval.

\textbf{Training objective.} We train the joint denoiser with flow matching~\cite{flowmatching}. 
For each demonstration window, we sample $\tau\sim\mathcal{U}(0,1)$ and select rolling or initialization mode with probabilities $\beta$ and $1-\beta$, respectively. 

Let $\boldsymbol{\sigma}_j$ denote the video and action noise levels of chunk $j$ under selected schedule. 
Let $\boldsymbol{\sigma}=(\boldsymbol{\sigma}_1,\ldots,\boldsymbol{\sigma}_W)$ denote the window noise profile. 
We construct the noisy targets:
\begin{equation}
    \widetilde{\mathbf{X}}_j
    =(1-\boldsymbol{\sigma}_j)\mathbf{X}_j
    +\boldsymbol{\sigma}_j\boldsymbol{\epsilon}_j,
    \qquad
    \boldsymbol{\epsilon}_j
    \sim\mathcal{N}(\mathbf{0},\mathbf{I}),
    \label{eq:flow_path}
\end{equation}
using independent Gaussian samples for video and actions.

The network predicts conditional velocities for all chunks, with target $\boldsymbol{\epsilon}_j-\mathbf{X}_j$. 
For modality $q\in\{v,a\}$, the per-chunk flow-matching loss is:
\begin{equation}
    \ell_j^q
    =
    \left\|
        \left[
            f_\theta^q\!\left(
                \widetilde{\mathbf{X}}_{1:W},
                \boldsymbol{\sigma};c_t
            \right)
        \right]_j
        - \left(
            \boldsymbol{\epsilon}_j^q-\mathbf{X}_j^q
        \right)
    \right\|_2^2,
    \label{eq:chunk_flow_loss}
\end{equation}
where $[\cdot]_j$ selects the prediction for chunk $j$, $\mathbf{X}_j^v=\mathbf{v}^{(j)}$, and $\mathbf{X}_j^a=\mathbf{a}^{(j)}$. 
The joint training objective is:
\begin{equation}
    \mathcal{L}
    =
    \mathbb{E}\!\left[
        \frac{1}{W}\sum_{j=1}^{W}
        b_j\,w(\boldsymbol{\sigma}_j)
        \left(\lambda_v\ell_j^v+\lambda_a\ell_j^a\right)
    \right],
    \label{eq:joint_objective}
\end{equation}
where $w$ weights the noise levels and $\lambda_v,\lambda_a$ balance the two modalities. 
The activity mask $b_j$ excludes initialization chunks held at pure noise. 
Training on both modes equips the same model to initialize the prediction window and continue its refinement during closed-loop execution.

\section{Experiments}

Our experiments verify whether rolling denoising can accelerate joint video-action prediction while preserving manipulation performance.
We compare task success with a broad set of vision-language-action models and world action models on LIBERO, RoboTwin 2.0, and real-world Unitree G1 tasks.
Controlled latency measurements quantify the computational savings of rolling denoising.
Ablations examine the roles of window size, training noise schedules, and cross-chunk action attention.

\subsection{Experimental Setup}
\label{sec:experimental_setup}
We evaluate Rolling-WAM on the following settings.
We report the task success rate (\%) as the primary performance metric and measure replanning latency to assess efficiency.

\subsubsection{LIBERO}
We evaluate the Spatial, Object, Goal, and Long suites of
LIBERO~\cite{libero}, each containing 10 tasks and 500
demonstrations. Observations include external and wrist RGB
views. We train one policy across all four suites for 10 epochs
with an effective batch size of 128, and evaluate each task
with 50 rollouts.

\subsubsection{RoboTwin 2.0}
We evaluate 50 bimanual manipulation tasks in the Clean and
Randomized settings of RoboTwin 2.0~\cite{robotwin}.
Observations include RGB images from the head and two wrist
cameras. We train one policy across 50 tasks using 2,500
clean demonstrations and 25,000 demonstrations from randomized
scenes, for 5 epochs with an effective batch size of 1024.
Each task is evaluated with 100 rollouts per setting.

\subsubsection{Real-world humanoid manipulation}
As shown in Fig.~\ref{fig:g1_real_world}, we evaluate three tasks on a Unitree G1 humanoid:
\emph{Doll Placement} (placing a plush dog in a box), 
\emph{Plate Stacking} (stacking three plates), and
\emph{Bead Pouring} (transferring beads from a bottle into a glass).
The policy receives a $320\times224$ egocentric RGB image 
and a 43-dimensional robot state. It predicts
78-dimensional actions comprising a 64-dimensional SONIC
motion latent~\cite{sonic} and 7-dimensional commands for each
hand. We train one policy across all three tasks using 50
demonstrations per task, recorded at 10\,Hz. Training runs for
7,500 steps with an effective batch size of 192. We execute
actions at 10\,Hz and evaluate 20 trials per task.

\subsection{Implementation Details}
We initialize the video expert from Wan2.2-TI2V-5B~\cite{wan2025}
and reuse its pretrained text encoder and VAE. The action
expert has 30 layers and hidden dimension $d_a=1024$
(approximately 1B parameters). Its backbone is initialized
by interpolating the video expert's weights~\cite{fastwam}.
We jointly train both experts and the proprioceptive encoder,
while freezing the VAE and text encoder. All training uses
benchmark demonstrations without additional embodied pretraining.

We use AdamW with learning rate $10^{-4}$, weight decay
$10^{-2}$, 5\% linear warmup followed by cosine decay, and
BF16 mixed precision. Video and action losses are averaged
separately, with $\lambda_v=\lambda_a=1$. Initialization and
rolling modes are sampled with probabilities 0.2 and 0.8.
Both modalities use the same shifted noise schedule~\cite{sd3-sigma-shift},
$\sigma(\tau)=\rho\tau/[1+(\rho-1)\tau]$ with $\rho=5$,
during training and inference.

Unless otherwise specified, we use $N=10$ denoising steps per
chunk, a window of $W=5$ chunks, and $K=16$ actions per chunk.
Each steady-state replanning cycle therefore uses $N/W=2$
denoising steps. The classifier-free guidance scale is 1.

\begin{table}[!t]
\caption{Success rates (\%) on LIBERO. P.T. indicates embodied pretraining. The highest and second-highest average scores are bold and underlined, respectively. \methodname achieves competitive performance with faster inference.
}
\label{tab:embodied_planning}
\centering
\small
\setlength{\tabcolsep}{0.8pt}
\renewcommand{\arraystretch}{1.08}
\begin{tabular*}{\columnwidth}{@{\extracolsep{\fill}}lcccccc@{}}
\toprule
{\footnotesize\bfseries Method} & \shortstack{\footnotesize\bfseries P.T.} & {\footnotesize\bfseries Spatial} & {\footnotesize\bfseries Object} & {\footnotesize\bfseries Goal} & {\footnotesize\bfseries Long} & {\footnotesize\bfseries Average} \\
\midrule
$\pi_{0}$~\cite{pi0}        & \cmark & 96.8 & 98.8  & 95.8 & 85.2 & 94.1 \\
$\pi_{0.5}$~\cite{pi05}     & \cmark & 98.8 & 98.2  & 98.0 & 92.4 & 96.9 \\
Motus~\cite{motus}          & \cmark & 96.8 & 99.8  & 96.6 & 97.6 & 97.7 \\
LingBot-VA~\cite{lingbotva} & \cmark & 98.5 & 99.6  & 97.2 & 98.5 & \textbf{98.5} \\
Fast-WAM~\cite{fastwam}     & \xmark & 98.2 & 100.0 & 97.0 & 95.2 & 97.6 \\
Joint-WAM~\cite{fastwam} & \xmark & 99.6 & 99.4  & 98.2 & 96.8 & \textbf{98.5} \\
\textbf{\methodname (Ours)} & \xmark & 98.2 & 98.0  & 98.2 & 97.8 & \underline{98.1} \\
\bottomrule
\end{tabular*}
\end{table}

\begin{table}[!t]
\caption{Success rates (\%) on RoboTwin 2.0. P.T. indicates embodied pretraining. The highest and second-highest average scores are bold and underlined, respectively. \methodname achieves the highest success rate in both settings.}
\label{tab:robotwin_main}
\centering
\small
\setlength{\tabcolsep}{3pt}
\renewcommand{\arraystretch}{1.08}
\begin{tabular*}{\columnwidth}{@{\extracolsep{\fill}}lcccc@{}}
\toprule
\textbf{Method} & \textbf{P.T.} & \textbf{Clean} & \textbf{Rand.} & \textbf{Average} \\
\midrule
$\pi_{0}$~\cite{pi0} & \cmark & 65.9 & 58.4 & 62.2 \\
$\pi_{0.5}$~\cite{pi05} & \cmark & 82.7 & 76.8 & 79.8 \\
Motus~\cite{motus} & \cmark & 88.7 & 87.0 & 87.8 \\
LingBot-VA~\cite{lingbotva} & \cmark & 92.9 & 91.5 & \underline{92.2} \\
Fast-WAM~\cite{fastwam} & \xmark & 91.9 & 91.8 & 91.8 \\
Joint-WAM~\cite{fastwam} & \xmark & 90.8 & 90.3 & 90.6 \\
\textbf{\methodname (Ours)} & \xmark & 93.5 & 93.0 & \textbf{93.3} \\
\bottomrule
\end{tabular*}
\end{table}

\textbf{Baselines.}
We compare with both VLA and WAM policies, including $\pi_{0}$~\cite{pi0}, $\pi_{0.5}$~\cite{pi05},
GR00T N1.7~\cite{gr00tn1_2025}, Motus~\cite{motus},
LingBot-VA~\cite{lingbotva}, Fast-WAM~\cite{fastwam}, and
Joint-WAM~\cite{fastwam}.
For controlled WAM comparisons, we reproduce
Fast-WAM and Joint-WAM using matched
training and evaluation settings wherever applicable.
Joint-WAM jointly denoises future video and actions from
Gaussian noise at each replanning cycle, while Fast-WAM generates
actions without future-video denoising at deployment.

\subsection{Performance on Simulation Benchmarks}
\label{sec:simulation_benchmarks}

\begin{figure}[t]
    \centering
    \includegraphics[width=\columnwidth]{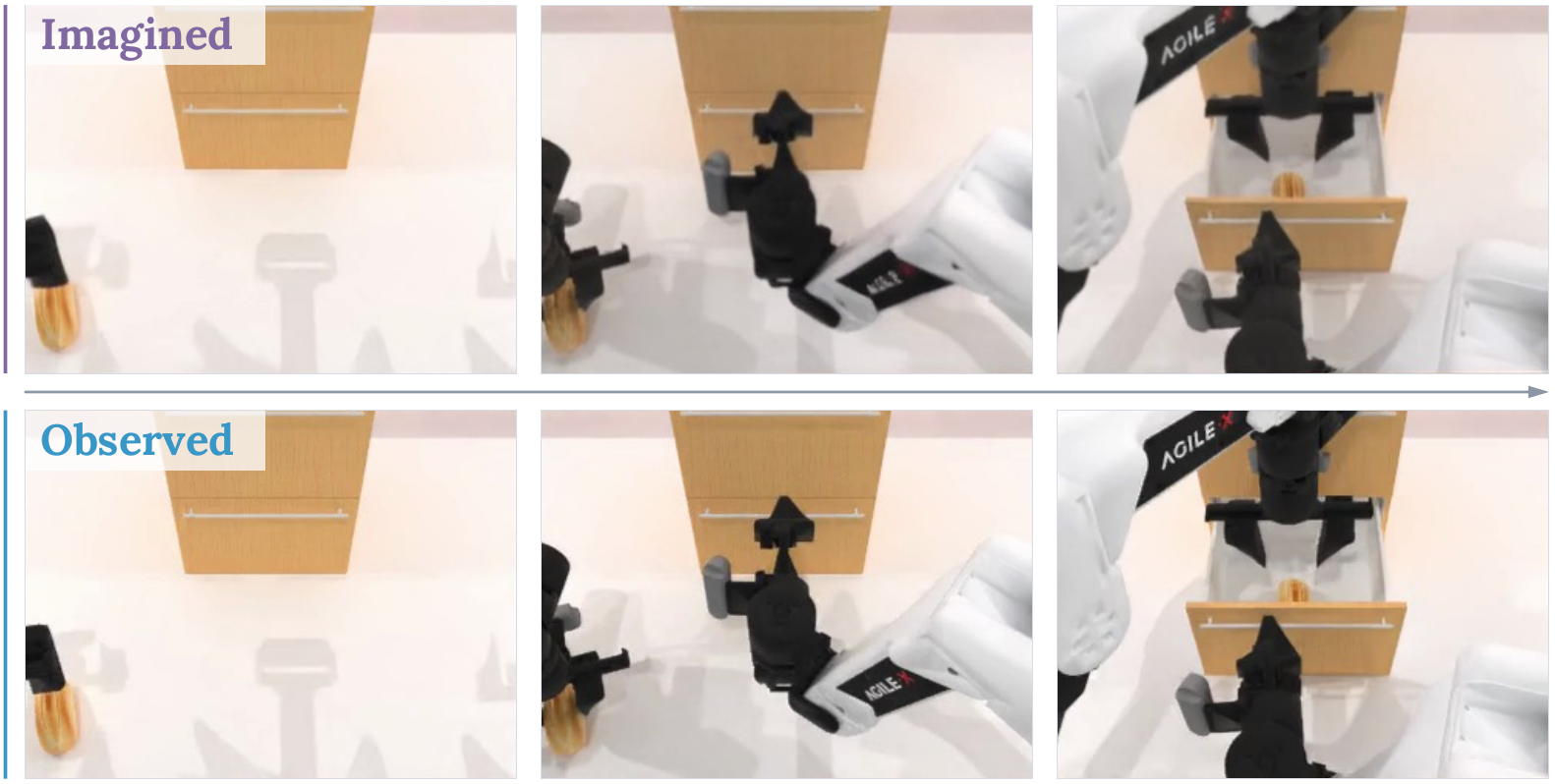}
    \caption{\textbf{Imagined and observed rollouts.}
    \methodname's imagined video (top) and ground-truth observations (bottom)
    during policy execution on \emph{Put Object Cabinet}.
    Columns show successive moments.}
    \vspace{-12pt}
    \label{fig:robotwin-imagined-observed}
\end{figure}

We first evaluate whether rolling denoising can deliver competitive
manipulation performance with fewer denoising steps per
replanning cycle.

As shown in Table~\ref{tab:embodied_planning}, \methodname
achieves an average success rate of 98.1\% on LIBERO, within 0.4 percentage points of the joint leaders, LingBot-VA and Joint-WAM (both 98.5\%), and above Motus (97.7\%) and Fast-WAM (97.6\%). Success ranges from 97.8\% to 98.2\% across the four suites,
showing consistently strong performance with rolling inference.

RoboTwin 2.0 further tests bimanual manipulation under scene
randomization. As shown in Table~\ref{tab:robotwin_main}, \methodname
achieves success rates of 93.5\% and 93.0\% in clean and
randomized settings, respectively. Its 93.3\% average leads LingBot-VA
(92.2\%), Fast-WAM (91.8\%), and Joint-WAM (90.6\%).
The high success rate in both settings shows that rolling
inference remains effective under the evaluated scene variations,
without additional embodied pretraining.
See Appendix Table~\ref{tab:robotwin_full_results} for per-task results.

\methodname achieves these results with only two denoising
steps per replanning cycle,
suggesting that co-refining current and future
predictions can potentially combine faster replanning with improved
manipulation performance.

We also inspect whether the retained visual predictions remain
aligned with observations as execution proceeds.
Figure~\ref{fig:robotwin-imagined-observed} compares imagined
and observed rollouts on \emph{Put Object Cabinet} in the RoboTwin 2.0 benchmark.
Imagined robot motion and task progression closely follow
the observations in this example, providing a qualitative view
of the visual context maintained during rolling execution.



\subsection{Inference Efficiency}
\label{sec:inference_efficiency}

To formally investigate the inference efficiency gains of
rolling denoising, we quantitatively compare \methodname
with the baselines and examine how replanning latency
varies with the prediction horizon.
We measure steady-state replanning latency on a single NVIDIA 
A100 GPU under the
RoboTwin 2.0 setting at an image resolution of $384\times320$.
All three methods execute 16 actions per replanning cycle.
Timing includes visual encoding and denoising, with CUDA
synchronization, and excludes warm-up and initialization.

\begin{figure}[t]
    \centering
    \includegraphics[width=0.95\columnwidth]{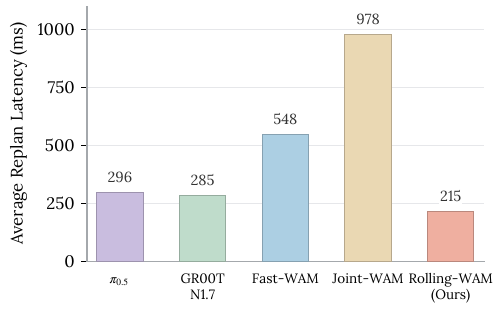}
    \vspace{-12pt}
    \caption{\textbf{Replanning latency comparison.}
    Average latency over multiple steady-state replanning cycles,
    excluding initialization. Lower is better.}
    \vspace{-12pt}
    \label{fig:robotwin-replan-latency}
\end{figure}

\textbf{Default configuration.}
As shown in Fig.~\ref{fig:robotwin-replan-latency}, with $N=10$ and $W=5$, \methodname
uses two denoising steps
per update and takes 215\,ms, compared with 978\,ms for
Joint-WAM and 548\,ms for Fast-WAM. These correspond to
approximately $4.5\times$ and $2.5\times$ speedups, respectively.
The VLA baselines $\pi_{0.5}$ and GR00T N1.7 take 296\,ms and 285\,ms per update, respectively.
\methodname thus retains future-video generation while taking
less time per update than either VLA baseline in this comparison.
In the meantime, \methodname retains an 80-action prediction window, while the
baselines predict 16 actions. The reduction in latency is
therefore achieved while refining a longer future at each update.
These timings use no \texttt{torch.compile}, TensorRT, or
custom CUDA kernels.

\textbf{Window-size scaling.} To examine how distributing denoising over more cycles affects latency, we vary the window size $W$ from 1 to 8. To match execution horizon, \methodname uses $H=16W$, while the baselines use $H=16$.
At each $W$, all methods use $N=W\lfloor16/W\rfloor$ total denoising
steps. Rolling inference uses $N/W=\lfloor16/W\rfloor$ steps per replanning cycle. We set the denoising step as 16 to keep rounding effects small across the tested window sizes.

\begin{figure}[t]
    \centering
    \includegraphics[width=0.95\columnwidth]{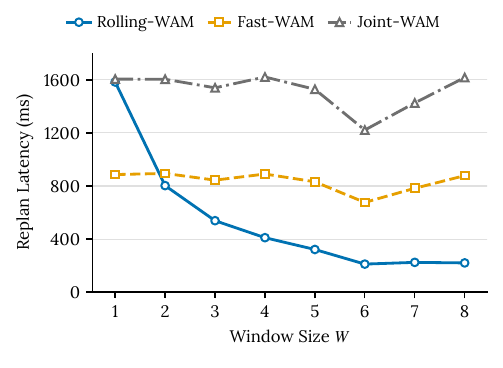}
    \vspace{-12pt}
    \caption{\textbf{Replanning latency across window sizes.}
    At each window size $W$, Rolling-WAM and the baselines use prediction horizons $H=16W$ and $H=16$, respectively, with the total denoising budget $N=W\lfloor 16/W\rfloor$. Points show mean replanning latency (ms), excluding initialization.}
    \label{fig:replan-latency}
\end{figure}

As shown in Fig.~\ref{fig:replan-latency}, at $W=5$, Rolling-WAM requires only 322\,ms, compared with 832\,ms for Fast-WAM and 1529\,ms for Joint-WAM,
yielding $2.58\times$ and $4.75\times$ speedups, respectively.
The latency curve flattens for $W=6$--8, where the number of
rolling denoising steps stays at two per replanning cycle. We use $W=5$ by default to balance latency and task performance (Sec.~\ref{sec:ablations}).


\begin{figure}[!t]
    \centering
    \includegraphics[width=\columnwidth]{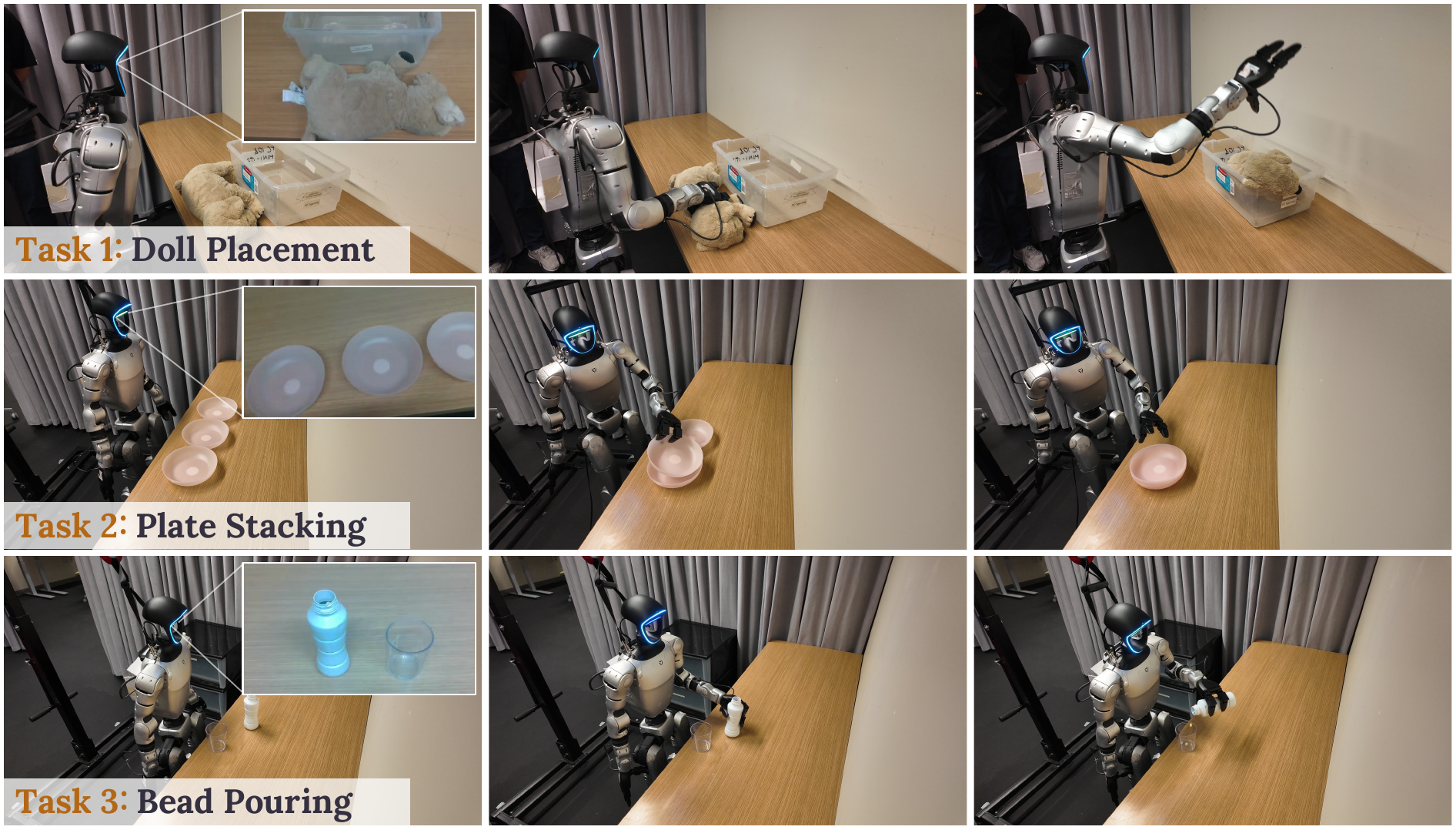}
    \caption{\textbf{Real-world humanoid manipulation tasks.}
    Doll Placement, Plate Stacking, and Bead Pouring (top to bottom),
    illustrated with qualitative examples from \methodname rollouts.
    Each row shows task progression from left to right. Insets show egocentric observations.}
    \label{fig:g1_real_world}
    \vspace{-12pt}
\end{figure}

\subsection{Real-World Humanoid Evaluation}

To assess the practical benefits of rolling denoising in
physical manipulation, we evaluate \methodname on real-world 
humanoid tasks. A trial succeeds when the
robot completes the task without human intervention: placing
the doll in the box, stacking all three plates, or transferring
beads into the glass. Figure~\ref{fig:g1_real_world} shows
representative rollouts.

As shown in Table~\ref{tab:g1_manipulation}, \methodname
achieves the highest average success rate of 85.0\% among
the five policies, followed by
Joint-WAM at 78.3\% and Fast-WAM at 75.0\%.
It matches the strongest baselines on \emph{Doll Placement} (85\%)
and \emph{Plate Stacking} (100\%), while achieving the highest success
rate on \emph{Bead Pouring} (70\%).
These results are obtained with a single policy trained
across all three tasks.

Qualitatively, we observe more continuous execution across
action-chunk boundaries with \methodname. Pauses are more
apparent with Joint-WAM and sometimes interrupt task progress.
These observations illustrate the practical importance of
replanning latency during physical manipulation.


\begin{table}[t]
\caption{Humanoid manipulation on Unitree G1. Each task is evaluated over 20 rollouts per method. Task cells report success rates (\%). The highest and second-highest average scores are bold and underlined, respectively.}
\label{tab:g1_manipulation}
\centering
\small
\setlength{\tabcolsep}{1.5pt}
\renewcommand{\arraystretch}{1.08}
\begin{tabular*}{\columnwidth}{@{\extracolsep{\fill}}lcccc@{}}
\toprule
\textbf{Method} & \shortstack{\textbf{Doll}\\\textbf{Placement}} & \shortstack{\textbf{Plate}\\\textbf{Stacking}} & \shortstack{\textbf{Bead}\\\textbf{Pouring}} & \shortstack{\textbf{Avg.}\\(\%)} \\
\midrule
$\pi_{0.5}$~\cite{pi05} & \gOnePiDog & \gOnePiPlate & \gOnePiPour & \gOnePiAverage \\
GR00T N1.7~\cite{gr00tn1_2025} & \gOneGrootDog & \gOneGrootPlate & \gOneGrootPour & \gOneGrootAverage \\
Fast-WAM~\cite{fastwam} & \gOneFastDog & \gOneFastPlate & \gOneFastPour & \gOneFastAverage \\
Joint-WAM~\cite{fastwam} & \gOneJointDog & \gOneJointPlate & \gOneJointPour & \underline{\gOneJointAverage} \\
\textbf{\methodname (Ours)} & \gOneRollingDog & \gOneRollingPlate & \gOneRollingPour & \textbf{\gOneRollingAverage} \\
\bottomrule
\end{tabular*}
\end{table}

\subsection{Ablation Studies}
\label{sec:ablations}

\begin{figure}[t]
    \centering
    \includegraphics[width=0.95\columnwidth]{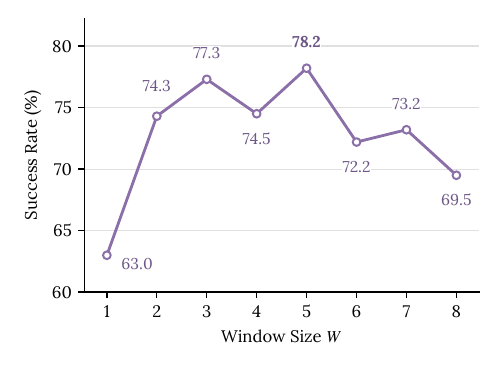}
    \vspace{-12pt}
    \caption{\textbf{Window size and task success.}
    Average success rates on the six selected RoboTwin tasks
    under the Clean setting. The prediction horizon is $H=16W$.}
    \vspace{-12pt}
    \label{fig:window-size-ablation}
\end{figure}

\begin{table}[t]
\caption{Ablations on noise scheduling and action attention. Mean success rates (\%) on six RoboTwin tasks and four LIBERO suites. A2A denotes bidirectional action attention across chunks; bold indicates the best per column.}

\label{tab:schedule_attention_ablation}
\centering
\small
\setlength{\tabcolsep}{3pt}
\renewcommand{\arraystretch}{1.08}
\begin{tabular*}{\columnwidth}{@{\extracolsep{\fill}}lcc@{}}
\toprule
\textbf{Variant} & \textbf{Selected RoboTwin} & \textbf{LIBERO} \\
\midrule
\textbf{Full Rolling (w/o A2A)} & 78.2 & \textbf{98.1} \\
w/ A2A & 76.3 & 97.9 \\
Constant ($p=0.2$) & 74.7 & 97.9 \\
Constant ($p=0.5$) & 73.0 & 97.1 \\
Random ($p=0.2$) & 75.3 & 97.0 \\
Random ($p=0.5$) & \textbf{78.5} & 97.3 \\
\bottomrule
\end{tabular*}

\end{table}

To support our design choices, we examine the prediction window, 
training noise schedules,
and action attention. Table~\ref{tab:schedule_attention_ablation}
reports results on LIBERO and six representative RoboTwin Clean tasks:
\emph{Lift Pot}, \emph{Beat Block Hammer}, \emph{Place Dual Shoes},
\emph{Stack Bowls Two}, \emph{Blocks Ranking Size}, and
\emph{Stack Blocks Three}.

\textbf{Window size.}
Increasing $W$ provides a longer predicted future, but its value
depends on whether the additional context helps the next action.
To test this, we vary $W=1,\ldots,8$ on the six selected RoboTwin
tasks, with $H=16W$ and $N=W\lfloor16/W\rfloor$, following the
sampling rule in Sec.~\ref{sec:inference_efficiency}.
As shown in Fig.~\ref{fig:window-size-ablation}, success peaks at 78.2\% for $W=5$, versus 77.3\% for $W=3$, and 69.5\% for $W=8$. Larger windows do not consistently
improve performance. We conjecture that more distant visual predictions are less
constrained by the current observation and may offer limited
guidance for the imminent action chunk. The best result in this
sweep comes from a moderate window, which motivates our choice
of $W=5$ as the default.

\textbf{Training noise mixture.}
Our default training matches the staggered noise profiles used
at inference. We test whether exposing the model to a broader
mixture improves policy performance.
We keep initialization training unchanged and replace a
fraction $p$ of rolling-mode samples with alternative noise
profiles. Constant uses one shared noise level across chunks;
Random samples a separate level for each chunk.
The results are shown in Table~\ref{tab:schedule_attention_ablation}. The full rolling schedule gives 78.2\% success on selected RoboTwin
tasks and 98.1\% on LIBERO.
Random with $p=0.5$ raises RoboTwin success to 78.5\%, but lowers
LIBERO to 97.3\%.
No alternative improves both benchmarks.
The full rolling schedule achieves the best LIBERO score and
is within 0.3 percentage points of the best RoboTwin variant.
We therefore retain the rolling schedule for all
non-initialization samples.

\textbf{Cross-chunk action attention.}
Action chunks already receive context from the entire visual
window. We test whether adding direct attention between action
chunks further improves performance.
As shown in Table~\ref{tab:schedule_attention_ablation}, allowing this attention gives 76.3\%
success on the selected RoboTwin tasks and 97.9\% on LIBERO,
compared with 78.2\% and 98.1\% with within-chunk action attention.
These results favor sharing information through the visual window
while restricting direct action attention to each chunk.

\begin{table}[t]
\caption{Compatibility with Faster-WAM~\cite{fasterwam}. Mean success rates (\%) and replanning latency (ms). Bold marks the best value per column.}
\label{tab:efficiency_compatibility}
\centering
\small
\setlength{\tabcolsep}{3pt}
\renewcommand{\arraystretch}{1.08}
\begin{tabular*}{\columnwidth}{@{\extracolsep{\fill}}lccc@{}}
\toprule
\textbf{Method} & \shortstack{\textbf{Selected}\\\textbf{RoboTwin}} & \textbf{LIBERO} & \shortstack{\textbf{Latency}\\\textbf{(ms)} $\downarrow$} \\
\midrule
Rolling-WAM & \textbf{78.2} & 98.1 & 215 \\
Faster-WAM~\cite{fasterwam} & 65.0 & \textbf{98.6} & 427 \\
Rolling-WAM-Faster & 75.0 & 97.5 & \textbf{142} \\
 \bottomrule
\end{tabular*}
\end{table}

\textbf{Combining with other efficiency methods.}
To explore the compatibility of \methodname with other efficiency methods,
we further combine it with the Faster-WAM
architecture~\cite{fasterwam}, which we denote as Rolling-WAM-Faster.
As shown in Table~\ref{tab:efficiency_compatibility}, 
under the same measurement protocol as in
Sec.~\ref{sec:inference_efficiency}, replanning latency
drops to 142\,ms, compared with 427\,ms for the original
Faster-WAM. The combined model also retains strong task performance.
These results demonstrate the compatibility of \methodname
with complementary architectural optimizations.

\section{Conclusion}
In this paper, we presented \methodname, an efficient World
Action Model that jointly refines near-term and future
video-action predictions within a sliding window.
By distributing denoising across replanning cycles, it
reduces the sequential steps needed to produce the next
executable action chunk, while using partially denoised
visual futures to guide action generation.
Experiments show that \methodname achieves competitive
performance on LIBERO and outperforms the compared methods
on RoboTwin 2.0 and real-world humanoid manipulation tasks, while delivering substantial speedups in inference over standard joint WAMs.
    Combining rolling denoising with complementary model and
system optimizations offers further opportunities for
low-latency robot control.

\textit{Limitations and future work.}
Design choices such as window and chunk sizes warrant further
exploration across task settings with different dynamics and
control frequencies. Despite observation feedback, retained
predictions may lag behind rapid scene changes and misguide
action generation, particularly with long windows.
Adaptive window management and asynchronous execution offer
directions for improving closed-loop responsiveness.




\section*{Acknowledgments}
We are grateful to our friends and colleagues, including Songlin Wei, Jiawei Yang, and Hongyi Jing, for their helpful discussions and insightful suggestions.
This work was partially supported by the National Science Foundation through NSF CPS \#2434460. The USC Physical Superintelligence Lab acknowledges generous support from Toyota Research Institute, Dolby, Google DeepMind, Capital One, Nvidia, Bosch, NSF, and Qualcomm. Yue Wang is also supported by a Powell Research Award.

\bibliographystyle{IEEEtran}
\bibliography{ref}

\appendix[RoboTwin Detailed Results]
Table~\ref{tab:robotwin_full_results} presents per-task success rates
for \methodname and the baselines on all 50 RoboTwin 2.0 tasks
under clean and randomized evaluation settings.

\begin{table*}[!t]
\caption{Per-task success rates (\%) on RoboTwin 2.0 under clean and randomized settings. Bold denotes the best score in each setting.}
\label{tab:robotwin_full_results}
\centering
\footnotesize
\setlength{\tabcolsep}{2.4pt}
\renewcommand{\arraystretch}{1.05}
\newcolumntype{C}{>{\centering\arraybackslash}X}
\begin{tabularx}{\textwidth}{lCC|CC|CC|CC|CC|CC}
\toprule
Task & \multicolumn{2}{c}{\textbf{Rolling-WAM}} & \multicolumn{2}{c}{Fast-WAM} & \multicolumn{2}{c}{Joint-WAM} & \multicolumn{2}{c}{LingBot-VA} & \multicolumn{2}{c}{$\pi_{0.5}$} & \multicolumn{2}{c}{Motus} \\
\cmidrule(lr){2-3}\cmidrule(lr){4-5}\cmidrule(lr){6-7}\cmidrule(lr){8-9}\cmidrule(lr){10-11}\cmidrule(lr){12-13}
 & Clean & Rand. & Clean & Rand. & Clean & Rand. & Clean & Rand. & Clean & Rand. & Clean & Rand. \\
\midrule
Adjust Bottle & 99 & 97 & \textbf{100} & \textbf{100} & 98 & 99 & 90 & 94 & \textbf{100} & 99 & 89 & 93 \\
Beat Block Hammer & 91 & 91 & 99 & 97 & \textbf{100} & \textbf{98} & 96 & \textbf{98} & 96 & 93 & 95 & 88 \\
Blocks Ranking RGB & 97 & 95 & \textbf{100} & \textbf{100} & \textbf{100} & \textbf{100} & 99 & 98 & 92 & 85 & 99 & 97 \\
Blocks Ranking Size & 83 & 87 & \textbf{94} & \textbf{98} & 83 & 91 & \textbf{94} & 96 & 49 & 26 & 75 & 63 \\
Click Alarmclock & \textbf{100} & 99 & \textbf{100} & \textbf{100} & \textbf{100} & \textbf{100} & 99 & \textbf{100} & 98 & 89 & \textbf{100} & \textbf{100} \\
Click Bell & \textbf{100} & \textbf{100} & \textbf{100} & \textbf{100} & \textbf{100} & 98 & \textbf{100} & \textbf{100} & 99 & 66 & \textbf{100} & \textbf{100} \\
Dump Bin Bigbin & 95 & 96 & \textbf{97} & 96 & 95 & 95 & 89 & 96 & 92 & \textbf{97} & 95 & 91 \\
Grab Roller & \textbf{100} & \textbf{100} & \textbf{100} & \textbf{100} & \textbf{100} & \textbf{100} & \textbf{100} & \textbf{100} & \textbf{100} & \textbf{100} & \textbf{100} & \textbf{100} \\
Handover Block & 94 & \textbf{93} & 95 & 81 & 93 & 91 & \textbf{99} & 78 & 66 & 57 & 86 & 73 \\
Handover Mic & 95 & \textbf{100} & 99 & \textbf{100} & \textbf{100} & \textbf{100} & 94 & 96 & 98 & 97 & 78 & 63 \\
Hanging Mug & 70 & 58 & 58 & \textbf{62} & \textbf{71} & 56 & 40 & 28 & 18 & 17 & 38 & 38 \\
Lift Pot & \textbf{100} & \textbf{100} & \textbf{100} & \textbf{100} & \textbf{100} & \textbf{100} & \textbf{100} & 99 & 96 & 85 & 96 & 99 \\
Move Can Pot & \textbf{98} & 97 & 90 & 88 & 97 & \textbf{99} & 94 & 97 & 51 & 55 & 34 & 74 \\
Move Pillbottle Pad & \textbf{100} & 99 & \textbf{100} & 99 & 99 & \textbf{100} & 99 & 99 & 84 & 61 & 93 & 96 \\
Move Playingcard Away & \textbf{100} & 97 & \textbf{100} & \textbf{100} & \textbf{100} & \textbf{100} & \textbf{100} & 99 & 96 & 84 & \textbf{100} & 96 \\
Move Stapler Pad & 83 & 84 & 77 & 64 & 85 & 81 & \textbf{91} & 79 & 56 & 42 & 83 & \textbf{85} \\
Open Laptop & 97 & 98 & \textbf{98} & \textbf{100} & 89 & 92 & 92 & 94 & 90 & 96 & 95 & 91 \\
Open Microwave & \textbf{100} & \textbf{91} & 62 & 45 & 3 & 14 & 82 & 86 & 34 & 77 & 95 & \textbf{91} \\
Pick Diverse Bottles & 85 & 87 & 80 & 85 & 86 & 87 & 89 & 82 & 81 & 71 & \textbf{90} & \textbf{91} \\
Pick Dual Bottles & 98 & \textbf{99} & \textbf{100} & 96 & 98 & \textbf{99} & \textbf{100} & \textbf{99} & 93 & 63 & 96 & 90 \\
Place A2B Left & \textbf{97} & 92 & 95 & 93 & 96 & \textbf{96} & \textbf{97} & 93 & 87 & 82 & 88 & 79 \\
Place A2B Right & \textbf{98} & 97 & 93 & \textbf{99} & 95 & 95 & 97 & 95 & 87 & 84 & 91 & 87 \\
Place Bread Basket & \textbf{97} & 93 & 91 & 93 & 89 & 94 & \textbf{97} & \textbf{95} & 77 & 64 & 91 & 94 \\
Place Bread Skillet & 93 & \textbf{98} & 90 & 93 & 90 & 93 & \textbf{95} & 90 & 85 & 66 & 86 & 83 \\
Place Burger Fries & 97 & 98 & 96 & 99 & \textbf{100} & \textbf{100} & 97 & 95 & 94 & 87 & 98 & 98 \\
Place Can Basket & 76 & 70 & 71 & 69 & 50 & 23 & \textbf{81} & \textbf{84} & 62 & 62 & \textbf{81} & 76 \\
Place Cans Plasticbox & \textbf{100} & \textbf{99} & 99 & 96 & 98 & 98 & \textbf{100} & \textbf{99} & 94 & 84 & 98 & 94 \\
Place Container Plate & 97 & 98 & 96 & \textbf{100} & \textbf{99} & 98 & \textbf{99} & 97 & \textbf{99} & 95 & 98 & 99 \\
Place Dual Shoes & 90 & \textbf{91} & \textbf{94} & 88 & 93 & 89 & \textbf{94} & 89 & 75 & 75 & 93 & 87 \\
Place Empty Cup & \textbf{100} & \textbf{100} & \textbf{100} & \textbf{100} & \textbf{100} & \textbf{100} & \textbf{100} & \textbf{100} & \textbf{100} & 99 & 99 & 98 \\
Place Fan & 93 & 93 & 96 & \textbf{96} & \textbf{99} & \textbf{96} & \textbf{99} & 93 & 87 & 85 & 91 & 87 \\
Place Mouse Pad & \textbf{97} & \textbf{98} & 83 & 89 & 96 & 91 & 93 & 96 & 60 & 39 & 66 & 68 \\
Place Object Basket & \textbf{95} & 86 & 89 & \textbf{88} & 86 & 81 & 91 & \textbf{88} & 80 & 76 & 81 & 87 \\
Place Object Scale & 91 & 98 & 90 & 97 & \textbf{96} & \textbf{99} & \textbf{96} & 95 & 86 & 80 & 88 & 85 \\
Place Object Stand & 90 & 93 & 90 & 94 & 92 & \textbf{98} & \textbf{99} & 96 & 91 & 85 & 98 & 97 \\
Place Phone Stand & \textbf{100} & 97 & 97 & 99 & \textbf{100} & \textbf{100} & 97 & 97 & 81 & 81 & 87 & 86 \\
Place Shoe & 95 & 98 & 96 & \textbf{99} & 95 & 97 & 98 & 98 & 92 & 93 & \textbf{99} & 97 \\
Press Stapler & 83 & 79 & 90 & 97 & 52 & 50 & 85 & 82 & 87 & 83 & \textbf{93} & \textbf{98} \\
Put Bottles Dustbin & \textbf{96} & \textbf{97} & 95 & 90 & 93 & 95 & 87 & 91 & 84 & 79 & 81 & 79 \\
Put Object Cabinet & 86 & \textbf{91} & 94 & 89 & \textbf{95} & 90 & 85 & 87 & 80 & 79 & 88 & 71 \\
Rotate QRcode & 88 & 78 & 93 & 89 & 91 & \textbf{92} & \textbf{96} & 91 & 89 & 87 & 89 & 73 \\
Scan Object & 95 & 90 & 89 & \textbf{92} & 92 & \textbf{92} & \textbf{96} & 91 & 72 & 65 & 67 & 66 \\
Shake Bottle & \textbf{100} & 99 & \textbf{100} & \textbf{100} & \textbf{100} & \textbf{100} & \textbf{100} & 97 & 99 & 97 & \textbf{100} & 97 \\
Shake Bottle Horizontally & \textbf{100} & \textbf{100} & \textbf{100} & \textbf{100} & \textbf{100} & \textbf{100} & \textbf{100} & 99 & 99 & 99 & \textbf{100} & 98 \\
Stack Blocks Three & \textbf{99} & \textbf{98} & 95 & 97 & 98 & 97 & \textbf{99} & \textbf{98} & 91 & 76 & 91 & 95 \\
Stack Blocks Two & \textbf{100} & \textbf{100} & \textbf{100} & \textbf{100} & \textbf{100} & \textbf{100} & \textbf{100} & 98 & 97 & \textbf{100} & \textbf{100} & 98 \\
Stack Bowls Three & 83 & 82 & 80 & 81 & 84 & 86 & \textbf{86} & 83 & 77 & 71 & 79 & \textbf{87} \\
Stack Bowls Two & 95 & 97 & 92 & \textbf{98} & 97 & 95 & 94 & \textbf{98} & 95 & 96 & \textbf{98} & \textbf{98} \\
Stamp Seal & \textbf{96} & 98 & 90 & 94 & \textbf{96} & \textbf{99} & \textbf{96} & 97 & 79 & 55 & 93 & 92 \\
Turn Switch & 65 & 73 & 61 & 59 & 73 & 72 & 44 & 45 & 62 & 54 & \textbf{84} & \textbf{78} \\
\midrule
\textbf{Average} & \textbf{93.54} & \textbf{92.98} & 91.88 & 91.78 & 90.84 & 90.32 & 92.90 & 91.50 & 82.74 & 76.76 & 88.66 & 87.02 \\
\bottomrule
\end{tabularx}
\end{table*}

\end{document}